\documentclass{llncs}  

\usepackage{graphicx} 
\usepackage[T1]{fontenc}
\usepackage{amsmath}
\usepackage[english]{babel}
\usepackage{array,tabularx}
\usepackage{booktabs} 
\usepackage{makeidx}  
\usepackage{xparse}
\usepackage{moredefs}
\usepackage{lips}
\usepackage{epstopdf} 
\usepackage{color}
\usepackage{csquotes}
\usepackage{xcolor}
\usepackage{times}
\usepackage[hidelinks]{hyperref}
\usepackage[nolist,nohyperlinks]{acronym}
\usepackage{comment}
\usepackage{paralist}
\usepackage{cleveref}
\usepackage{wrapfig}
\usepackage{multirow}
\usepackage{microtype}
\usepackage{listings}
\usepackage[super]{nth}
\usepackage{natbib}
\usepackage{booktabs}
\usepackage{siunitx}
\usepackage{threeparttable}
\usepackage{tabulary}
\usepackage{amsfonts}
\usepackage{subcaption}

\newtheorem{hypothesis}{H\kern-0.2em}
\makeatletter
\newcounter{parenthypothesis}
\newenvironment{subhypothesis}
 {
  \refstepcounter{hypothesis}%
  \protected@edef\theparenthypothesis{\thehypothesis}%
  \setcounter{parenthypothesis}{\value{hypothesis}}%
  \setcounter{hypothesis}{0}%
  \def\thehypothesis{\theparenthypothesis\alph{hypothesis}}%
  \ignorespaces
}{%
  \setcounter{hypothesis}{\value{parenthypothesis}}%
  \ignorespacesafterend
}
\makeatother

\usepackage{float}
\floatstyle{plaintop}
\restylefloat{table}

\usepackage[style=base,labelfont=bf,labelsep=period,tableposition=top,font=small]{caption}
\makeatletter
\renewcommand\@biblabel[1]{#1.}
\makeatother
\addto\captionsenglish{}

\usepackage{fancyhdr}
\fancypagestyle{WI_footer}{\fancyhf{}\fancyfoot[L]{\color{black}\scriptsize 19th International Conference on Wirtschaftsinformatik\\September 2024, Würzburg, Germany}}

\crefformat{footnote}{#2\footnotemark[#1]#3}
\expandafter\def\expandafter\UrlBreaks\expandafter{\UrlBreaks
  \do\a\do\b\do\c\do\d\do\e\do\f\do\g\do\h\do\i\do\j%
  \do\k\do\l\do\m\do\n\do\o\do\p\do\q\do\r\do\s\do\t%
  \do\u\do\v\do\w\do\x\do\y\do\z\do\A\do\B\do\C\do\D%
  \do\E\do\F\do\G\do\H\do\I\do\J\do\K\do\L\do\M\do\N%
  \do\O\do\P\do\Q\do\R\do\S\do\T\do\U\do\V\do\W\do\X%
  \do\Y\do\Z}

\newcolumntype{L}[1]{>{\raggedright\arraybackslash}p{#1}}   
\newcolumntype{C}[1]{>{\centering\arraybackslash}p{#1}}     
\newcolumntype{R}[1]{>{\raggedleft\arraybackslash}p{#1}}    

\begin{document}
\frontmatter          

\mainmatter              

\title{The Impact of Transparency in AI Systems on Users’ Data-Sharing Intentions: A Scenario-Based Experiment}

\subtitle{Research Paper} 
\author{Julian Rosenberger\inst{1} \and
Sophie Kuhlemann\inst{2} \and
Verena Tiefenbeck\inst{2} \and
Mathias Kraus\inst{1} \and
Patrick Zschech\inst{3}}

\institute{
Universität Regensburg, Bajuwarenstraße 4, 93053 Regensburg\\
\email{{julian.rosenberger, mathias.kraus}@ur.de}  \and
Friedrich-Alexander Universität Erlangen-Nürnberg, Lange Gasse 20, 90403 Nürnberg\\
\email{{sophie.kuhlemann, verena.tiefenbeck}@fau.de} \and
Universität Leipzig, Grimmaische Straße 12, 04109 Leipzig\\
\email{patrick.zschech@uni-leipzig.de}
}

\maketitle
\setcounter{footnote}{0}

\begin{abstract}
Artificial Intelligence (AI) systems are frequently employed in online services to provide personalized experiences to users based on large collections of data. However, AI systems can be designed in different ways, with black-box AI systems appearing as complex data-processing engines and white-box AI systems appearing as fully transparent data-processors. As such, it is reasonable to assume that these different design choices also affect user perception and thus their willingness to share data. To this end, we conducted a pre-registered, scenario-based online experiment with 240 participants and investigated how transparent and non-transparent data-processing entities influenced data-sharing intentions. Surprisingly, our results revealed no significant difference in willingness to share data across entities, challenging the notion that transparency increases data-sharing willingness. Furthermore, we found that a general attitude of trust towards AI has a significant positive influence, especially in the transparent AI condition, whereas privacy concerns did not significantly affect data-sharing decisions.\\ 

\end{abstract}

\thispagestyle{WI_footer}


\section{Introduction}
\label{sec:introduction}
Artificial Intelligence (AI) systems are increasingly used in online services to provide a personalized experience to users. However, many AI systems lack transparency regarding how for example individual recommendations are derived, making them appear as black-boxes to users \citep{herlocker2000explaining, millecamp2019explain, sinha2002role}. This lack of transparency may promote user aversion to AI systems, leading to a preference for human involvement in certain cases \citep{cadario2021understanding}, despite AI's demonstrated superiority in many tasks \citep{brown2019superhuman, esteva2017dermatologist}. Research in the fields of Explainable Artificial Intelligence (XAI) and Interpretable Machine Learning (IML) aims to tackle this aversion towards algorithms by providing insights into the inner workings of AI systems \citep{rudin2019stop, adadi2018peeking}. While XAI tries to provide a glimpse into complex black-box models, IML focuses on developing and applying simpler, clearer models for inherent transparency \citep{poursabzi2021manipulating}.
Both fields intend to display the outcomes of an AI system in a way that humans can understand. Prior research already suggests that "opening the black-box" \citep{mahmud2022influences} by providing explanations of the decision-making processes of an AI system can positively affect users' willingness to engage with the system \citep{yeomans2019making, shin2021effects, herlocker2000explaining, millecamp2019explain, dzindolet2003role}. However, research shows that factors such as trustworthiness and familiarity of a system influence users' reliance on both, human advisors and AI systems \citep{choung2023trust, chellappa2005personalization, lee2004trust, rotter1967new}. 
 
Despite the reliance of AI systems on user data to provide a personalized experience, there has been little research on how XAI or IML affect users' attitudes on privacy \citep{rai2020explainable}. Privacy studies have shown that when users are faced with the decision to share their data with a service provider, they tend to weigh the costs and benefits of disclosing their data \citep{dinev2006privacy}. 
For instance, it shows that people are more willing to share their personal data if they can see a clear benefit, such as receiving personalized offers \citep{white2004consumer}. Transparent systems that demonstrate how sharing specific types of personal data can result in more personalized and valuable services thus might have the potential to encourage consumers to share their data for improved experiences. However, providing detailed explanations of data-processing may also make privacy concerns salient to consumers, which could make them less willing to share their personal information \citep{kehr2015blissfully}. Several studies show that individuals' privacy concerns significantly impact their risk assessment of data-sharing, which ultimately affects their willingness to share their data \citep{cichy2021privacy, dinev2006extended, malhotra2004internet, son2008internet, li2011role}. 

This study aims to examine the impact of transparent (white-box AI) and non-transparent (black-box AI) data-processing systems on users' intentions to share data. Building on the insights of prior research suggesting a preference for human decision-making over algorithms \citep{cadario2021understanding, longoni2019resistance}, we include a human data-processor as a comparative element. Additionally, we are examining how trust factors such as the perceived trustworthiness of different data-processing systems affects users' willingness to share data, based on the research by \citet{chellappa2005personalization} and \citet{choung2023trust}. Lastly, we assess the impact of users' privacy concerns within this dynamic. Consequently, we pose the following research questions:

\vspace{6pt}
\noindent\textbf{RQ} \, \textit{How does users' willingness to share personal data vary between different data-processing entities, specifically a human expert, a white-box AI system, and a black-box AI system? Furthermore, how do trust in the entity and privacy concerns affect the relationship between data-processing entity and the intention to disclose data?} 
\vspace{6pt}

To answer our research questions, we conducted a scenario-based experiment in which we presented participants (N = 240) with a fictitious sleep app that aimed to provide tailored advice based on the data entered by the user. Our study contributes to existing AI research on users' adoption of interpretable systems, as well as to privacy research, particularly with respect to data-sharing behavior. In the following section, Section \ref{sec:related_work}, we discuss related work from (X)AI and IML research and its connection to trust and privacy concerns and derive our research model. Section \ref{sec:methods} summarizes the experimental design followed in our study. In Section \ref{sec:results}, we present the results of our study. Section \ref{sec:discussion} discusses our findings and concludes our work. 

\section{Conceptual Background and Related Work}
\label{sec:related_work}

Many AI systems are characterized by their inability to convey human-interpretable information about how and why they make certain decisions \citep{herlocker2000explaining, millecamp2019explain, sinha2002role}, referred to as black-box AI systems \citep{du2019techniques, rudin2019stop, adadi2018peeking}. This lack of transparency into the system's decision-making processes can cause people to be hesitant in adopting AI outputs. For example, in the context of healthcare, individuals were shown to feel more confident in understanding the choices made by human healthcare professionals compared to ones made by algorithms. As a result, they typically favor human providers over algorithmic ones \citep{cadario2021understanding, longoni2019resistance}, even though AI systems have been proven to perform better than humans in various tasks \citep{esteva2017dermatologist, brown2019superhuman}. Additionally, if a human agent is framed as an expert, users are even more likely to depend on human rather than on AI advice: This was demonstrated by \citet{madhavan2007effects} in a study in which they examined users' trust in and perceived reliability of human and automated systems based on written descriptions, distinguishing between novices and experts. 
Users' tendency to rather rely on human advice than on AI decision-making is commonly known as algorithm aversion and is often attributed to users' desire for perfect predictions and their low tolerance for errors made by an algorithm \citep{dietvorst2015algorithm, dietvorst2018overcoming}. However, recent evidence suggests that there are situations in which human actors prefer AI advice, such as the prediction of business or geopolitical events \citep{logg2019algorithm}. Furthermore, providing explanations for errors made by a system has been found to positively influence users' perception of the system \citep{dzindolet2003role}. 

Along that line, research in the area of XAI and IML intends to address the potential reluctance towards algorithms by offering insight into the underlying reasoning of a system \citep{poursabzi2021manipulating}. Presenting AI systems in an understandable manner has the potential to mitigate the problems associated with their black-box nature while ensuring high predictive accuracy \citep{rudin2019stop, adadi2018peeking}. In recent years, a significant effort within the AI research community has thus been the development of XAI methods to make black-box AI systems more understandable \citep{ribeiro2016should, lundberg2017unified, koh2017understanding}. Despite these advances, concerns persist regarding the adequacy of these explanations for capturing the full complexity of the systems \citep{rudin2019stop, stiglic2020interpretability, babic2021beware}. To address the inherent approximations of XAI methods, white-box AI systems attempt to provide complete transparency and reliable decision-making through intrinsic interpretability \citep{du2019techniques}. For example, a white-box AI system for diagnosing a specific medical condition might use a transparent decision tree that consists of a series of interpretable if-then rules, allowing for clear tracing of the decision-making process. In contrast, a black-box AI system might employ a complex neural network comprising multiple layers that learn intricate relationships between features, their interactions, and the diagnosis. This multi-layer structure enables the neural network to capture non-linear patterns and high-order interactions, but at the cost of interpretability \citep{tjoa2020survey}. However, white-box AI system systems typically face limitations, most notably that they are primarily limited to tabular data \citep{kraus2023interpretable}. Numerous studies indicate that showing users clarifications can positively impact their perception of the system \citep{yeomans2019making, shin2021effects, herlocker2000explaining, millecamp2019explain, dzindolet2003role}. Nevertheless, \citet{alufaisan2021does} and \citet{schrills2020color} found no such positive effect on users' intention when provided with visual explanations. Previous research has also shown that when humans collaborate with AI, providing explanations can cause users to trust the system too much, leading to "automation bias" \citep{schemmer2022influence}. This can result in users accepting incorrect suggestions from the system \citep{bansal2021does}. In this context, a study by \citet{choung2023trust} shows that factors, such as the trustworthiness of a system and trust in its functionality, significantly influence users' intention to use a system. Similarly, the results of \citet{chellappa2005personalization} indicate that familiarity and previous experience, play a significant role in the use of personalization services. These findings are consistent with research on interpersonal trust, which argues that attitudes of trust play an important role in how people rely on each other \citep{lee2004trust, rotter1967new}.
 
Notably, despite the reliance of AI systems on user data, there has been little research on the impact of providing insights to an AI system on the privacy-related behavior of users \citep{rai2020explainable}. Information systems research has explored users' attitudes toward privacy in various contexts such as e-commerce \citep{acquisti2005privacy, spiekermann2001privacy}, financial services \citep{norberg2007privacy}, social networks \citep{barnes2006privacy}, and smartphone applications \citep{egelman2013choice}, focusing on users' attitudes towards privacy and its impact on their willingness to participate in e-commerce services \citep{chellappa2005personalization, eastlick2006understanding, pavlou2007understanding}, as well as their willingness to share personal information with these services \citep{belanger2002trustworthiness, dinev2006extended}. Studies in the context of users' data-sharing behavior suggest that while people consider protecting their personal data important, they often act against their preferences and willingly disclose their data (see \citet{gerber2018explaining} and \citet{kokolakis2017privacy} for reviews). This contradiction between privacy attitudes and behaviors has been studied extensively and is commonly known as the privacy paradox \citep{norberg2007privacy}. Studies revealed that users weigh their sharing intentions based on a cost-benefit analysis of data disclosure \citep{culnan1993did, dinev2006extended, dinev2006privacy, xu2009role}. Within this cost-benefit analysis, described as privacy calculus, one of the benefits already studied is personalization \citep{awad2006personalization}. In this context, individuals are found to be more willing to disclose personal information if they receive customized offers in response \citep{white2004consumer}. However, \citet{kehr2015blissfully} demonstrates that the tendency to focus on perceived benefits can be reduced by making privacy considerations salient to users by reminding them about risks associated with sharing sensitive data. Numerous studies show that information privacy concerns significantly impact individuals' risk assessment towards information privacy and thus affect their willingness to disclose their data \citep{cichy2021privacy, dinev2006extended, malhotra2004internet, son2008internet, li2011role}.

\subsection{Research Model}
\label{sec:research_model}

\begin{figure}[htp]
    \centering    
    \includegraphics[width=0.7\textwidth]{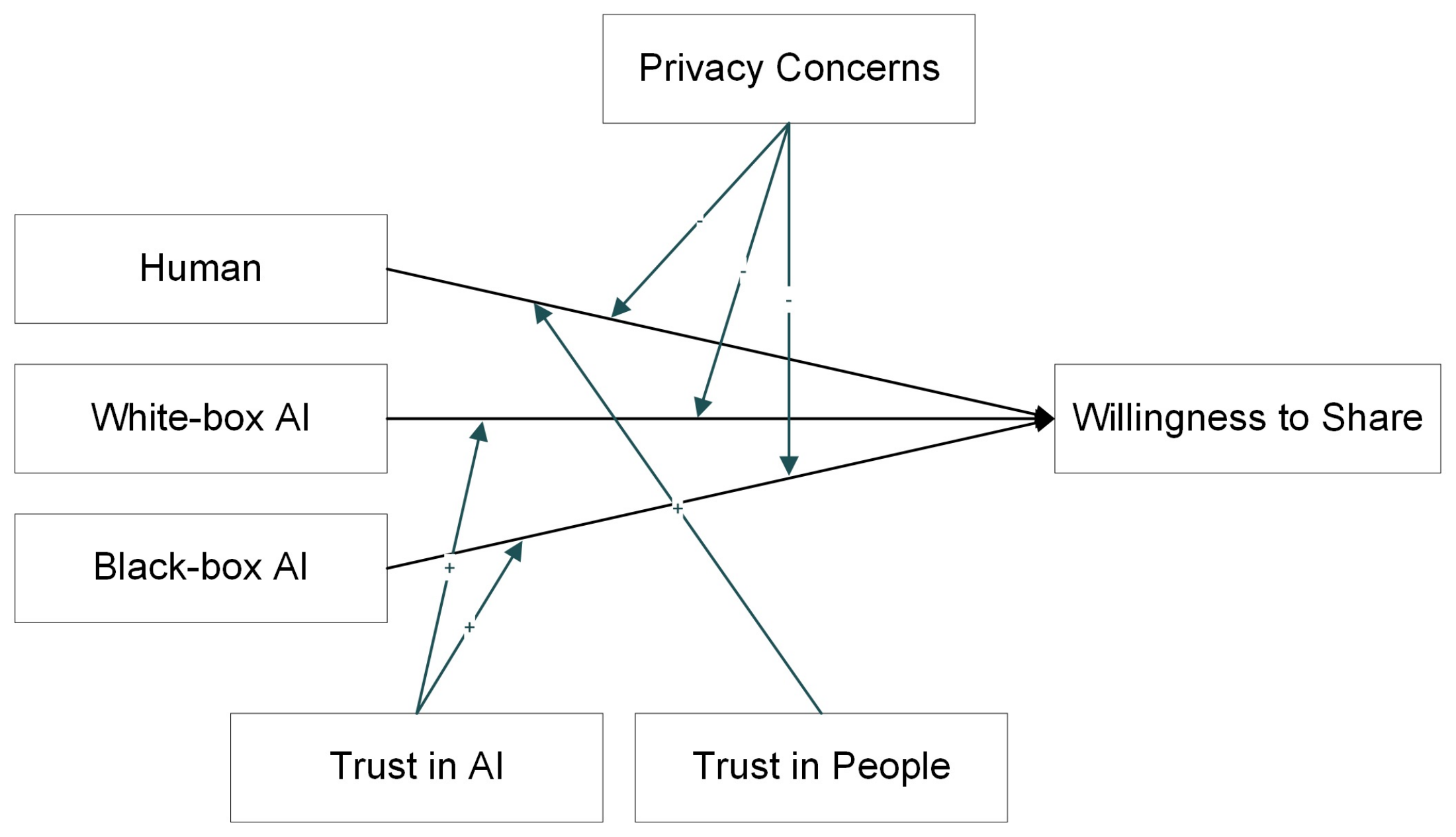}
    \caption{Research Model}
    \label{fig:research_model}
\end{figure}

More in-depth research is needed to investigate the impact of a system's data-processing nature on users' cost-benefit trade-offs when disclosing data, and whether consumers' willingness to share their data changes as a result \citep{rai2020explainable}. Systems that effectively communicate how sharing specific types of personal data can result in more personalized and valuable services may have the potential to motivate consumers to share their data for enhanced experiences \citep{awad2006personalization, white2004consumer}. However, transparent systems that provide detailed explanations of data-processing may also raise privacy concerns among consumers, leading them to be less willing to share their personal data \citep{kehr2015blissfully}. We thus build a research model (see Figure \ref{fig:research_model}) that, based on previous literature on the "black-box problem" \citep{herlocker2000explaining, millecamp2019explain, sinha2002role} and algorithm aversion \citep{dietvorst2015algorithm, dietvorst2018overcoming}, distinguishes between human, white-box AI and black-box AI data-processing entities and assesses their impact on users' sharing intentions.


As previous research has indicated that users tend to prefer human service providers over algorithmic ones \citep{cadario2021understanding, longoni2019resistance}, it is reasonable to assume that this notion is also reflected in the privacy-intentions of users as they may perceive human handlers of their personal data as more trustworthy compared to AI systems. Based on this, we hypothesize the following:

\begin{subhypothesis}
\label{hyp:entity}
\begin{hypothesis}
\label{hyp:human}
    Users exhibit greater willingness to share personal data when the data is processed by a human entity compared to AI entities.
\end{hypothesis}

However, white-box AI processing, as suggested by \citet{rudin2019stop}, could potentially serve as a mitigator to the concerns related to the black-box nature of common AI systems. Based on this assumption, we derive the following hypothesis in the context of users' data-sharing behavior:

\begin{hypothesis}
\label{hyp:AI}
    Users exhibit greater willingness to share personal data when the data is processed by a white-box AI entity compared to a black-box AI entity.
\end{hypothesis}
\end{subhypothesis}

Additionally, prior work suggests that factors such as the trustworthiness of a system and trust in its functionality significantly determine users' intention to use a system \citep{chellappa2005personalization, choung2023trust, lee2004trust}. We assume that the intention to engage with a system is closely related to user data-sharing, as interactions with AI systems often involve sharing personal data (e.g., sharing location data with map apps or entering health information into a symptom management app). Thus, we expect trust to have a moderating effect on the relationship between AI system and data-sharing willingness:

\begin{subhypothesis}
\label{hyp:trust}
\begin{hypothesis}
\label{hyp:trust_AI}
    Trust in AI will moderate the relationship between the AI groups and users' willingness to share personal data. Specifically, when trust in AI is high, participants in the white-box AI and black-box AI conditions will show a higher willingness to share data.
\end{hypothesis}

Likewise, we assume that the same moderating effect occurs for the human condition, as literature on interpersonal trust indicates that trust plays a relevant role in how much people rely on each other \citep{lee2004trust, rotter1967new}. Thus, we hypothesize:

\begin{hypothesis}
\label{hyp:trust_people}
    Trust in people will moderate the relationship between the human group and users' willingness to share personal data. Specifically, when trust in people is high, participants in the human condition will show a higher willingness to share data.
\end{hypothesis}
\end{subhypothesis}

Finally, we build on research by \citet{dinev2006extended, malhotra2004internet, son2008internet} and \citet{li2011role} suggesting that privacy concerns are a relevant determinant in users' privacy-related decision-making and argue that the relationship between entity and willingness to disclose data is influenced by the individual's concern about their privacy:

\begin{hypothesis}
\label{hyp:privacy_concerns}
    Privacy concerns will moderate the relationship between the human, white-box AI, and black-box AI groups and users' willingness to share sensitive health-related data. Specifically, when privacy concerns are high, participants in all conditions will show a lower willingness to share data.
\end{hypothesis}

\section{Methods}
\label{sec:methods}

\subsection{Experimental Design}
\label{sec:experimental-design}

We tested our hypotheses by conducting an online experiment based on a pre-registered scenario-based experiment with 240 participants.\footnote{Pre-registration link: \href{https://doi.org/10.17605/OSF.IO/U6FN7}{https://doi.org/10.17605/OSF.IO/U6FN7}. The experimental procedure received ethical clearance by the German Association for Experimental Economic Research (No. o3RhQZKR).} For our study, we chose an environment where subjects could quickly familiarize themselves with the topic and understand the underlying relationships without requiring in-depth knowledge. Specifically, our experiment is a use case related to healthy living and well-being, in which we refer to the dynamics between health and lifestyle factors and sleep.  Participants were introduced to a mock-up of SleepOptima, a hypothetical sleep app that aims to provide tailored advice based on user data input. To assess the effects of different data-processing entities on users' data-sharing behavior, we randomly assigned participants to one of three treatment groups representative of the different entities: a human expert, a white-box AI system, and a black-box AI system.

Each condition was presented with identically structured text modules containing detailed explanations of the data-processing techniques used by the respective entity within SleepOptima. In the human condition, it was clarified based on \citet{madhavan2007effects} that the expert's recommendation is derived from their acquired knowledge. The white-box and black-box conditions were identical in communicating information about the relevant system and how advice is generated for participants. However, the two groups differed in their clarification of the system's transparency, with the interpretability of the white-box AI being presented in contrast to the lack of interpretability of the black-box AI. To emphasize the different analytical approaches of the individual data-processing entities, we also provided visual aids to participants. For the human group, we used an image of a medical expert to emphasize the human element in data-processing. Based on \citet{du2019techniques} and \citet{rudin2019stop}, we chose a decision tree to symbolize the transparent and structured nature of the white-box AI. In contrast, for the black-box AI, a neural network illustration was used to depict the complexity and opacity of its processes. For a detailed presentation of the textual and visual treatments, refer to our online appendix.\footnote{Link to online appendix: \href{https://doi.org/10.17605/OSF.IO/8W6G3}{https://doi.org/10.17605/OSF.IO/8W6G3}}

\subsection{Measurement and Scales}
\label{sec:measurement}

To capture participants' willingness to share their data, we used a 7-point Likert scale -- from 1 ("Strongly Unwilling") to 7 ("Strongly Willing") -- across seven data categories relevant to the SleepOptima app. The categories range from demographics, over physical and mental status until questions about sexual activities. Each category consisted of two items and by calculating the average scores of these 14 items, we derived a continuous measure of each participant's willingness to share sensitive health-related data. In addition, we measured trust in the processing entity and privacy concerns using validated scales that we slightly adapted to our context.

\textit{Trust in AI} was assessed using an adapted scale from \citet{hoffman2023measures}, tailored to the context of our study comparing human and AI data-processing entities. This adaptation expanded the original scale to a 7-point Likert scale to capture a broad range of Trust in AI and to align it with other scales in our questionnaire. The statements included in this scale explore general trust in AI systems, perceived reliability, and predictability of AI outcomes.

\textit{Trust in People} was measured by adapting the General Trust Scale from \citet{yamagishi1994trust} to ensure consistency with other measures. This scale assesses people's general trustworthiness and honesty, exploring beliefs about people's basic honesty, trustworthiness, and goodness.

\textit{Privacy Concerns} were measured using the Internet Users' Information Privacy Concerns (IUIPC) scale developed by \citet{malhotra2004internet}, which provides an exploration of online Privacy Concerns. This scale was selected because of its alignment with the digital context of the study and its focus on consumer perspectives on information privacy \citep{gross2021validity}. Participants rated their agreement on a 7-point Likert scale, addressing issues such as discomfort with online companies requesting personal information and the importance of control over personal information.

By averaging scores across items for each construct, we derived continuous measures of Trust in AI, Trust in People, and Privacy Concerns in order to test our hypotheses. For further details on the scores for the different health-related categories and items, as well as on the operationalization of constructs, see our online appendix.

\subsection{Participants}
\label{sec:participants}

For our study, we recruited 331 participants over two days in October 2023 using Prolific. The study took approximately 6 minutes and 30 seconds to complete, and participants were compensated £14,17/hr for their time. To ensure data integrity, three attention checks were incorporated into the survey. The rate of failed attention checks was evenly distributed across all three conditions, and as a result, 91 participants were excluded based on this criterion. The final sample consisted of 240 participants, detailed in Table \ref{tab:demographics} with their demographic distribution. Our participant sample comprised 122 females (50.8\,\%), 114 males (47.5\,\%), one non-binary, and three who preferred not to disclose.

We conducted randomization checks to confirm that all demographic characteristics were equally distributed across the treatment groups. To assess the uniformity of this distribution, we used Fisher's exact test, which showed no significant differences in gender distribution across the groups $(p = 0.820)$. Additionally, we performed an Analysis of Variance (ANOVA), to examine age disparities. The results showed that there were no significant differences in the mean age between the groups $(F(2,\, 235) = 0.079,\, p = 0.924)$. These findings suggest that our sample had a balanced representation of demographics across groups.

\begin{table}[H]
    \centering
    \small
    \caption{Descriptive results of key socio-demographic data of the study sample across treatment groups (Human, White-box AI, Black-box AI).}
    \label{tab:demographics}
    \sisetup{
       separate-uncertainty,
       table-number-alignment=center
    }
    \begin{tabular*}{\textwidth}{
        @{}@{\extracolsep{\fill}}l
        S[table-format=2.0]
        S[table-format=2.1, table-space-text-post=\,\%]
        S[table-format=2.0]
        S[table-format=2.1, table-space-text-post=\,\%]
        S[table-format=2.0]
        S[table-format=2.1, table-space-text-post=\,\%]
        S[table-format=1.0]
        S[table-format=1.1, table-space-text-post=\,\%]
        S[table-format=2.1(2)]
    }
    \toprule
    & \multicolumn{2}{c}{Participants} & \multicolumn{6}{c}{Gender Identity} & {Age} \\
    \cmidrule(lr){4-9}
    & & & \multicolumn{2}{c}{Female} & \multicolumn{2}{c}{Male} & \multicolumn{2}{c}{Other$^\dagger$} & \\
    \midrule
    Human         & 83  & 34.6\,\% & 43 & 51.8\,\% & 37 & 44.6\,\% & 3 & 3.6\,\% & 40.3(15.9) \\
    White-box AI & 81 & 33.8\,\% & 38 & 46.9\,\% & 42 & 51.9\,\% & 1 & 1.2\,\% & 39.4(14.2) \\
    Black-box AI    & 76  & 31.7\,\% & 41 & 53.9\,\% & 35 & 46.1\,\% & 0 & 0.0\,\% & 40.1(15.0) \\
    \bottomrule
    \end{tabular*}
    \\[1ex]
    \parbox{\linewidth}{
        $^\dagger$ "Other" includes one non-binary participant and those who chose not to disclose their gender.
    }
\end{table}

\subsection{Manipulation Check}
\label{sec:manipulation}

To check if participants' perceptions aligned with our intended manipulations, we asked them to rate the transparency level of the received treatment on a 7-point Likert scale ranging from 1 "Low Level of Transparency" to 7 "High Level of Transparency" immediately after presenting the manipulation. This manipulation check was conducted before measuring the dependent variable (willingness to share data) to ensure that participants had accurately perceived the intended manipulation. The results showed that the human group $(M = 5.41,\, SD = 1.23)$ reported a notably higher level of transparency than the black-box AI group $(M = 3.92,\, SD = 2.11)$. The white-box AI group's perception $(M = 5.19,\, SD = 1.21)$ was aligned with the human group. An ANOVA confirmed these observations as statistically significant $(F(2,\, 237) = 20.67,\, p < .001)$. To examine these differences more closely, we conducted pairwise t-tests with a Bonferroni correction, revealing significant differences between the human and black-box AI groups $(p < .001$) and between the black-box AI and white-box AI groups $(p < .001)$, but no significant difference between the human and white-box AI groups. These results indicate that participants perceived distinct levels of transparency across the different data-processing entities, with the black-box AI being viewed as less transparent than both the human and white-box AI. 

\section{Results}
\label{sec:results}

\subsection{Overall Willingness to Share Data}
\label{sec:results_willingness}

Our findings revealed subtle differences in the willingness to share data across conditions. Participants showed the highest willingness to share data with the black-box AI system $(M = 6.02$, $SD = 0.85)$, slightly exceeding the willingness to share with a human expert $(M = 5.85$, $SD = 0.94)$ and the white-box AI system $(M = 5.82$, $SD = 1.04)$. We opted for regression analysis throughout the paper to quantify group differences, estimate effect sizes, and allow for the inclusion of interaction effects in subsequent analyses. The regression results, with the human group as a natural choice for comparison, revealed no significant differences between conditions (see Table \ref{tab:regression_willingness}). Thus, H1a, proposing greater willingness to share data with a human entity compared to AI entities, is not supported.

Further regression analysis testing H1b, which suggested greater willingness to share data with a white-box AI entity compared to a black-box AI entity, also revealed no significant differences between the two AI conditions $(F(1,\, 157) = 1.686,\, p = 0.196)$. Consequently, H1b cannot be confirmed based on our study's evidence.

\begin{table}[h!]
    \centering
    \scriptsize
    \caption{Simple regression result for Model Willingness to Share}
    \label{tab:regression_willingness}
    \begin{threeparttable}
    \sisetup{
       separate-uncertainty,
       table-number-alignment=center,
       table-space-text-post=\textsuperscript{***},
       print-zero-integer=true       
    }
    \begin{tabular*}{0.6\textwidth}{@{\extracolsep{\fill}}
        l
        S[table-format=-1.2(3)]
        S[table-format=<1.6]@{}}
        \toprule
        & \multicolumn{2}{c}{Model Willingness to Share} \\
        \cmidrule{2-3}
        {Variable} & {$B \pm SE$} & {$p$} \\
        \midrule
        \textbf{Intercept} & 5.85(0.11) & {<.001\textsuperscript{***}} \\[0.1cm]

        White-box AI & -0.03(0.15) & .867 \\
        Black-box AI & 0.17(0.15) & .258 \\
        \midrule
        $\text{\textit{R}}^2$ & \multicolumn{2}{c}{0.008} \\
        $\text{\textit{R}}^2_{\text{\textit{Adj}}}$ & \multicolumn{2}{c}{0.000} \\
        \bottomrule
    \end{tabular*}
    \begin{tablenotes}[flushleft]
        \scriptsize
        \item \textit{Note.} Ref = Human
    \end{tablenotes}
    \end{threeparttable}
\end{table}

\subsection{The Role of Trust in Data-Sharing Decisions}
\label{sec:results-trust}

We investigated the influence of trust on participants' willingness to disclose personal information across different data-processing entities. We used two regression models (Model Trust in AI for AI conditions and Model Trust in People for the human condition) and analyzed interactions through simple slopes analysis, following \citet{aiken1991multiple}.

\emph{Model Trust in AI.} Regression analysis (see Table \ref{tab:regression_trust}) shows that Trust in AI positively influences willingness to share data in both AI conditions, with a stronger effect in the white-box AI condition $(B = 0.49,\, p < .001)$ compared to the black-box AI condition $(B = 0.24,\, p = 0.08)$. Trust in AI had no effect in the human expert condition.

\begin{figure}[h!]
\centering
\includegraphics[width=\linewidth]{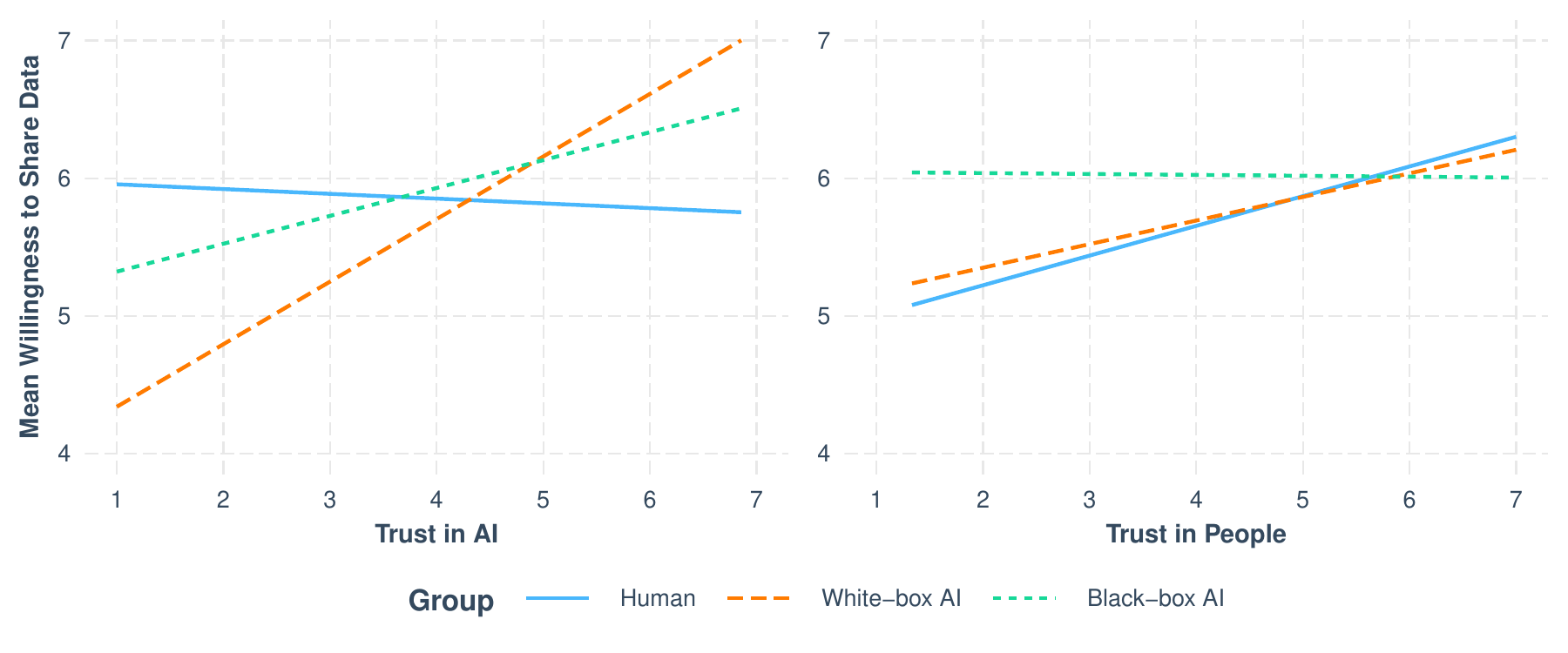}
\caption{Simple slopes plots illustrating the relationship between Trust in AI, Trust in People, and Mean Willingness to Share Data for Human, White-box AI, and Black-box AI groups.}
\label{fig:sim_slopes_trust}
\end{figure}

Simple slopes analysis (see Figure \ref{fig:sim_slopes_trust}) confirms these results, showing a modest but significant increase in willingness to share data for the black-box AI group $(B = 0.20,\, p = 0.04)$ and a more pronounced effect for the white-box AI condition $(B = 0.45,\, p < .001)$. The human condition showed negligible impact of Trust in AI $(B = -0.03,\, p = 0.71)$. Therefore, we can partially confirm H2a: Trust in AI has a significant positive effect on the willingness to share data in the white-box AI condition, as evidenced by the regression results, and a modest but significant positive effect in the black-box AI condition, as revealed by the simple slope analysis.

\emph{Model Trust in People.} The interaction effect of Trust in People $(B = 0.22,\, p = 0.054)$ is shown in Table \ref{tab:regression_trust}. Simple slopes analysis revealed additional information for the human group $(B = 0.22,\, p = 0.05)$ and the white-box AI group $(B = 0.17,\, p = 0.11)$. Based on these results, we cannot confirm H2b, as the data do not provide sufficient evidence to conclude that Trust in People significantly impacts data-sharing willingness in the human condition.

\begin{table}[h!]
    \centering
    \scriptsize
    \caption{Multiple regression results for Model Trust in AI and Model Trust in People}
    \label{tab:regression_trust}
    \begin{threeparttable}
    \sisetup{
       separate-uncertainty,
       table-number-alignment=center,
       table-space-text-post=\textsuperscript{***},
       print-zero-integer=true       
    }
        \begin{tabular*}{\textwidth}{@{\extracolsep{\fill}}
            l
            S[table-format=-1.2(3)]
            S[table-format=<1.6]
            S[table-format=-1.2(3)]
            S[table-format=<1.6]
            @{}}
            \toprule
            & \multicolumn{2}{c}{Model Trust in AI} & \multicolumn{2}{c}{Model Trust in People} \\
            \cmidrule{2-3} \cmidrule{4-5}
            {Variable} & {$B \pm SE$} & {$p$} & {$B \pm SE$} & {$p$} \\
            \midrule
            \textbf{Intercept} & 5.84(0.10) & {<.001\textsuperscript{***}} & 5.83(0.10) & {<.001\textsuperscript{***}} \\[0.1cm]
            White-box AI & -0.02(0.14) & 0.913 & 0.00(0.14) & 0.986 \\
            Black-box AI & 0.14(0.15) & 0.332 & 0.19(0.15) & 0.211 \\

            \textbf{Trust}\textsuperscript{\textdagger} \\
            Trust in AI & -0.03(0.09) & 0.712 & & \\
            Trust in People & & & 0.22(0.11) & 0.054 \\[0.1cm]

            \textbf{Group $\times$ Trust} \\
            White-box AI $\times$ Trust in AI & 0.49(0.13) & {<.001\textsuperscript{***}} \\
            Black-box AI $\times$ Trust in AI & 0.24(0.14) & 0.080 \\
            White-box AI $\times$ Trust in People & & & -0.04(0.15) & 0.773\\
            Black-box AI $\times$ Trust in People & & & -0.22(0.14) & 0.118 \\
            \midrule
            $\text{\textit{R}}^2$ & \multicolumn{2}{c}{0.113} & \multicolumn{2}{c}{0.035} \\
            $\text{\textit{R}}^2_{\text{\textit{Adj}}}$ & \multicolumn{2}{c}{0.094} & \multicolumn{2}{c}{0.014} \\
            \bottomrule
        \end{tabular*}
        \begin{tablenotes}[flushleft]
            \scriptsize
            \item \textit{Note.} \textsuperscript{\textdagger}Trust is mean-centered; Ref = Human
        \end{tablenotes}
    \end{threeparttable}
\end{table}

\subsection{Impact of Privacy Concerns on Willingness to Share Data}
\label{sec:results-privacy_concerns}

Regarding the stated moderating effect of Privacy Concerns on the relationship between data-processing entity and data-sharing intentions, the results of the regression analysis in Table \ref{tab:regression_privacy_concerns} show that Privacy Concerns alone did not significantly predict willingness to share in any condition. This finding suggests that Privacy Concerns did not negatively influence participants' data-sharing behavior, regardless of the data-processing entity involved. Therefore, we cannot confirm H3. 

\begin{table}[htb]
    \centering
    \scriptsize
    \caption{Multiple regression result for Model Privacy Concerns}
    \label{tab:regression_privacy_concerns}
    \begin{threeparttable}
    \sisetup{
       separate-uncertainty,
       table-number-alignment=center,
       table-space-text-post=\textsuperscript{***},
       print-zero-integer=true       
    }
        \begin{tabular*}{0.8\textwidth}{@{\extracolsep{\fill}}
            l
            S[table-format=-1.2(3)]
            S[table-format=<1.6]@{}}
            \toprule
            & \multicolumn{2}{c}{Model Privacy Concerns} \\
            \cmidrule{2-3}
            {Variable} & {$B \pm SE$} & {$p$} \\
            \midrule
            \textbf{Intercept} & 5.86(0.11) & {<.001\textsuperscript{***}} \\[0.1cm]

            White-box AI & -0.04(0.15) & 0.789 \\
            Black-box AI & 0.14(0.15) & 0.343 \\

            \textbf{Privacy Concerns}\textsuperscript{\textdagger} & -0.09(0.14) & 0.536 \\[0.1cm]

            \textbf{Group $\times$ Privacy Concerns} \\
            White-box AI $\times$ Privacy Concerns & -0.16(0.19) & 0.419 \\
            Black-box AI $\times$ Privacy Concerns & -0.03(0.18) & 0.886 \\
            \midrule
            $\text{\textit{R}}^2$ & \multicolumn{2}{c}{0.030} \\
            $\text{\textit{R}}^2_{\text{\textit{Adj}}}$ & \multicolumn{2}{c}{0.009} \\
            \bottomrule
        \end{tabular*}
        \begin{tablenotes}[flushleft]
            \scriptsize
            \item \textit{Note.} \textsuperscript{\textdagger}Mean-centered; Ref = Human
        \end{tablenotes}
    \end{threeparttable}
\end{table}

\section{Discussion}
\label{sec:discussion}

Our research explored the influence of different data-processing entities (human expert, white-box AI, and black-box AI) on individuals' willingness to share personal data, considering the moderating effects of trust and privacy concerns. Surprisingly, we found no significant differences in data-sharing willingness across the three conditions, contrary to previous findings on algorithm aversion \citep{dietvorst2015algorithm} and algorithm appreciation \citep{logg2019algorithm}. However, trust in AI emerged as a critical factor, significantly increasing data-sharing intentions in both white-box and black-box AI systems. Those results are aligned with the findings of \citet{choung2023trust} stating that factors such as trust in the functionality of an AI system affect users' intentions to use AI. Privacy concerns, on the other hand, did not significantly predict willingness to share data, thus are not in line with broad research previously conducted indicating that information privacy concerns impact individuals’ risk assessment towards information privacy and thus affect their willingness to disclose their data \citep{dinev2006extended, li2011role, malhotra2004internet, son2008internet} 

As such, our findings make several important theoretical and practical contributions to the literature on AI transparency, trust, and privacy. First, by showing no significant differences between treatment groups, our results challenge the prevailing academic discourse that argues that the inherent opacity of some AI systems can lead to problems such as a lack of user adoption which should be mitigated by building more transparent systems \citep{adadi2018peeking, poursabzi2021manipulating, rudin2019stop}. Indeed, our results emphasize that providing transparent systems alone is not enough to convince users of AI systems, but that creating trust in AI systems in general may also be beneficial. Finally, by examining users' willingness to share data as our key dependent variable, we provide initial insights into how the nature of the data-processing entity influences individuals' privacy calculus for mobile application services \citep{rai2020explainable}. This contribution enriches the academic discourse on privacy and trust in AI while offering practical guidance for developers on optimizing data-sharing in AI systems based on users' general trust in the system.


However, while our study provides valuable insights, it is important to acknowledge certain limitations. We tested our hypotheses using a mock-up of a fictitious sleep app, which may not fully replicate real-world user experiences. Future studies should explore data-sharing behaviors in more realistic settings. Additionally, our focus on purely human, white-box AI, and black-box AI systems does not account for hybrid systems common in real-world applications \citep{dellermann2019hybrid}. Future research should investigate user perceptions, trust, and interaction strategies for these hybrid systems. Another limitation is that we measured trust only after the experimental manipulation, preventing randomization checks for pre-existing trust levels. Future research should measure trust at multiple time points to control for potential confounding effects. Lastly, our study did not collect comprehensive demographic data (including education, income, profession, and geographical location) and other potentially influential factors like AI experience, literacy, and knowledge. Future research should consider including these variables to provide a more detailed understanding of data-sharing attitudes and behaviors.

In conclusion, our study makes significant contributions to the fields of AI and information privacy by examining the impact of data-processing entities on data-sharing intentions, considering the critical roles of trust and privacy concerns. Our findings offer valuable theoretical insights and practical recommendations for developing AI systems that prioritize transparency, trust, and user privacy. Future research should build upon our work by exploring data-sharing behaviors in different contexts, investigating the effects of hybrid AI systems, and examining the potential negative implications of excessive trust in AI \citep{greulich2024exploring}. By continuing to advance our understanding of these complex issues, we can develop AI systems that are not only technologically sophisticated but also ethically responsible and socially beneficial.
\newpage



\bibliographystyle{agsm}
\bibliography{_references}




\end{document}


\pagestyle{headings}  

\mainmatter     

\appendix
\label{appendix}

\section{Treatments}
\label{appendix:treatments}

\textbf{Human}

\begin{figure}[h!]
    \centering
    \includegraphics[width=0.67\textwidth]{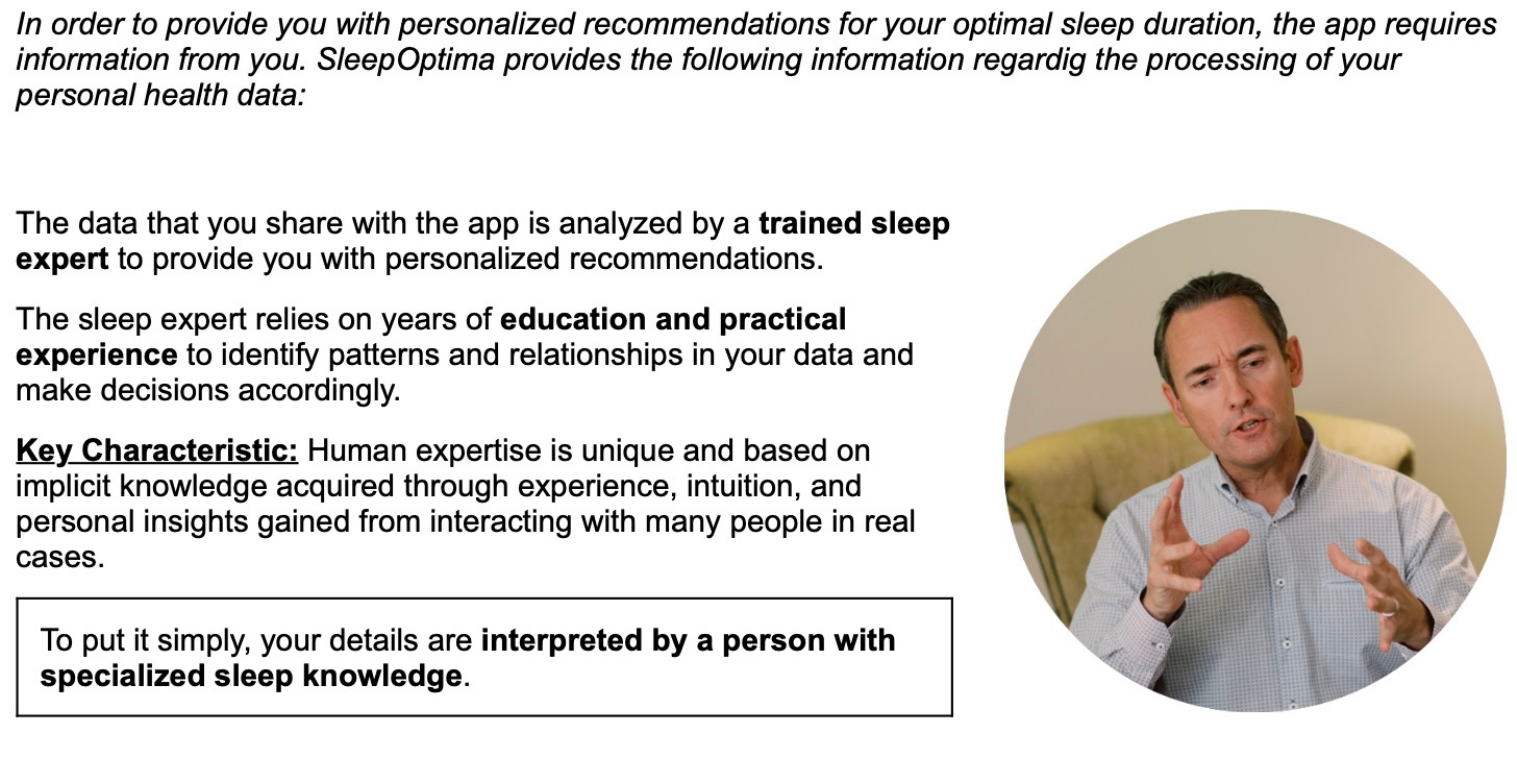}
    \caption{Visual and textual treatments for the Human condition}
    \label{fig:human_treatment}
\end{figure}

\noindent\textbf{White-box AI}

\begin{figure}[h!]
    \centering
    \includegraphics[width=0.67\textwidth]{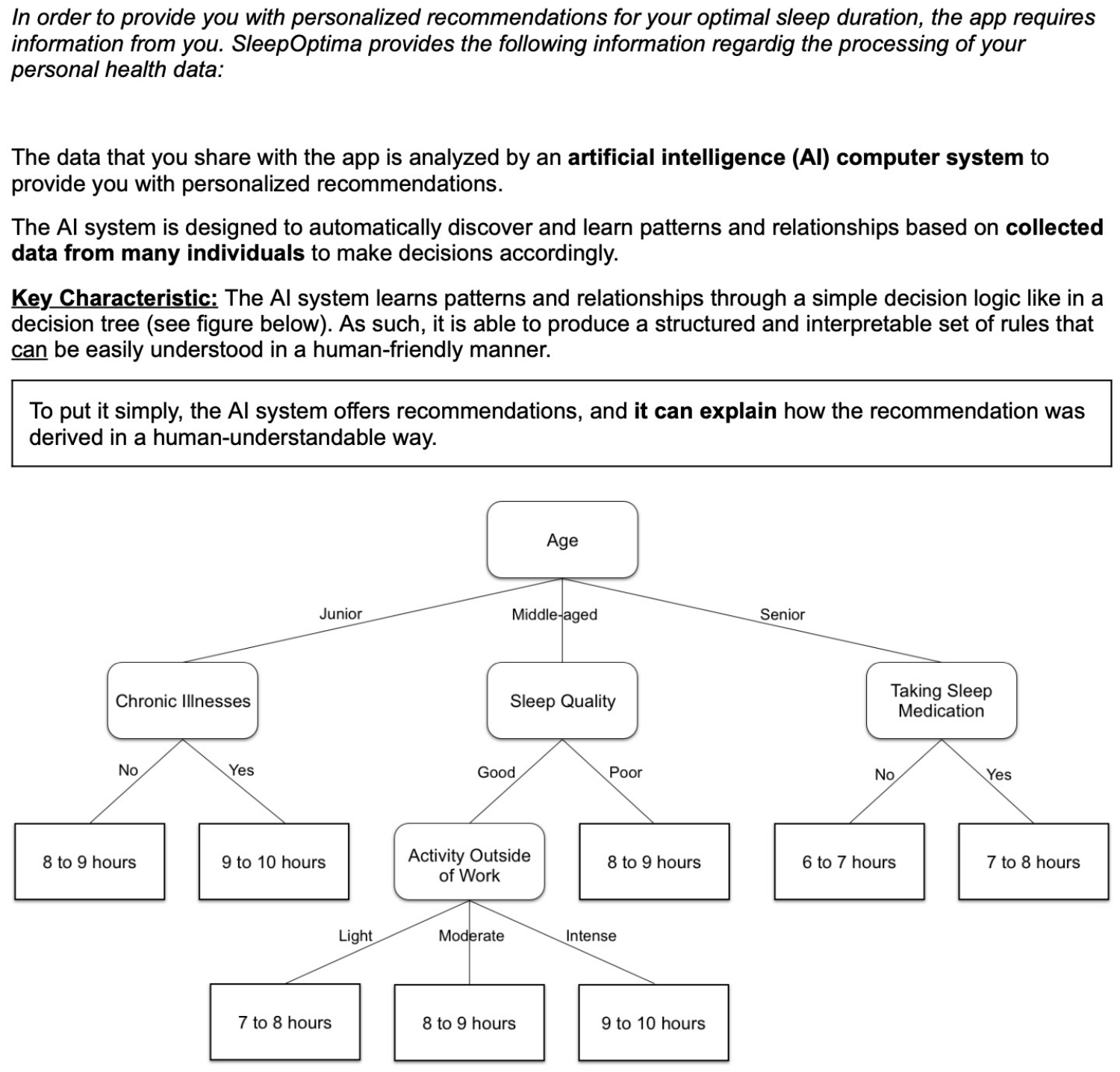}
    \caption{Visual and textual treatments for the White-box AI condition}
    \label{fig:white-box_treatment}
\end{figure}

\newpage
\noindent\textbf{Black-box AI}

\begin{figure}[h!]
    \centering
    \includegraphics[width=0.67\textwidth]{figures/Black-box Treatment.pdf}
    \caption{Visual and textual treatments for the Black-box AI condition}
    \label{fig:black-box_treatment}
\end{figure}

\section{Willingness to Share Data}
\label{app:willingness}

We assessed participants' data-sharing willingness for the SleepOptima app mock-up using a 14-item, 7-point Likert scale developed specifically for this study. The scale ranged from 1 ("Strongly Unwilling") to 7 ("Strongly Willing") and covered seven data categories, including demographics, physical and mental health, and sexual activities. Each category contained two items. We calculated the average of these 14 items to obtain a continuous measure of willingness to share sensitive health-related data, with mean scores and standard deviations categorized by treatment condition detailed in Table \ref{tab:willingness_comparison}.

Prior to the main study, we conducted a pre-test with $N = 30$ participants from Prolific to ensure that the questions were appropriately ordered based on their sensitivity level. The pre-test participants rated their willingness to share different types of information with health and well-being apps on the same 7-point Likert scale ("Strongly Unwilling" to "Strongly Willing"). The pre-test encompassed three items for each of the seven categories. Results from the pre-test showed that seven items had no clear differences in sensitivity and were thus removed from the survey. The remaining 14 items (two per category) were ordered according to the sensitivity classification derived from the pre-test results and included in our main survey (see Table \ref{tab:willingness_comparison}).

To assess the scale's reliability, we calculated Cronbach's alpha. The analysis revealed a raw alpha of 0.93 and a standardized alpha of 0.94, indicating high internal consistency reliability. However, it is important to note that a Cronbach's alpha greater than 0.9 may also suggest potential redundancy among the scale items \citep{streiner2003starting}. Further research is needed to examine the scale's dimensionality and determine whether some items could be measuring the same underlying construct.

The item-total correlations ranged from 0.61 to 0.86, suggesting that each item contributes positively to the overall scale reliability. However, given the high Cronbach's alpha, future studies should investigate the scale's factor structure to ensure that the items are not overly redundant.


\begin{table}[H]
    \centering
        \caption{Mean (M) and standard deviation (SD) of willingness to share personal data for the 14 requested items, differentiated by our three treatment conditions: Human expert, White-box AI system, and Black-box AI system.}
        \label{tab:willingness_comparison}
        
        \sisetup{
           separate-uncertainty,
           table-number-alignment=center,
           print-zero-integer=true
        }
        \begin{tabular*}{\textwidth}{@{\extracolsep{\fill}} l
            S[table-format=2.2(2)]
            S[table-format=2.2(2)]
            S[table-format=2.2(2)]
        }
        \toprule
        \textit{Category} & {Human} & {White-box AI} & {Black-box AI} \\
        \cmidrule{2-2} \cmidrule{3-3} \cmidrule{4-4}
        {Item} & {$M \pm SD$} & {$M \pm SD$} & {$M \pm SD$} \\
        \midrule
        \textit{Demographics} & & & \\
        Gender & 6.58(0.67) & 6.45(1.03) & 6.57(0.77)  \\
        Age & 6.57(0.69) & 6.48(0.90) & 6.61(0.67) \\[0.1cm]
        \textit{Activity Level} & & & \\
        Average Exercise Days per Week & 6.44(0.82) & 6.31(1.08) & 6.39(0.91) \\
        Activity Outside of Work & 6.42(0.85) & 6.31(1.09) & 6.39(0.78) \\[0.1cm]
        \textit{Sleep History} & & & \\
        Sleep Quality & 6.57(0.63) & 6.40(0.95) & 6.51(0.74) \\
        Sleep Medication & 6.20(1.04) & 6.27(1.03) & 6.34(1.14) \\[0.1cm]
        \textit{Physical Health Status} & & & \\
        Chronic Illness & 5.91(1.27) & 5.75(1.32) & 6.13(1.19) \\
        Chronic Medication & 5.85(1.32) & 5.69(1.37) & 5.97(1.30) \\[0.1cm]
        \textit{Mental Health Status} & & & \\
        Stress Level & 6.14(0.92) & 5.94(1.19) & 6.13(1.00) \\
        Mental Health Disorder & 5.53(1.57) & 5.60(1.45) & 5.74(1.27) \\[0.1cm]
        \textit{Substance Use} & & & \\
        Influence Substances Sleep & 5.48(1.72) & 5.72(1.51) & 5.86(1.35) \\
        Usage of Certain Substances & 5.14(1.89) & 5.63(1.71) & 5.72(1.49) \\[0.1cm]
        \textit{Sexual Activities} & & & \\
        Satisfaction Sex Life & 4.24(1.88) & 4.23(1.92) & 4.82(1.83) \\
        Sexual Medication & 4.80(2.05) & 4.75(1.93) & 5.08(1.79) \\
    \bottomrule
        \end{tabular*}
\end{table}

\section{Additional Constructs}
\label{appendix:additional_constructs}

\noindent \textbf{Trust in AI}\\

\noindent To operationalize the construct of Trust in AI, we adapted the trust scale for the XAI context from \cite{hoffman2023measures} to align with our investigation's focus on comparing the willingness to share data between human and AI data-processing entities. Recognizing the need for consistency across our measures and to increase the sensitivity of our analysis, we expanded the original 5-point Likert scale to a 7-point scale ranging from 1 ("Strongly Disagree") to 7 ("Strongly Agree"). In addition, we generalized the trust statements to ensure the applicability of the scale across all participant groups, including those without direct AI experience. This approach allowed us to capture a broad spectrum of Trust in AI, which is essential for analyzing its influence on user data-sharing behavior.

The adapted scale consists of seven items. By averaging the scores of these items, we obtain a continuous measure of Trust in AI.

\begin{enumerate}
    \item I generally feel confident in Artificial Intelligence systems. They tend to work well.
    \item The outputs from Artificial Intelligence systems are usually predictable.
    \item In general, Artificial Intelligence systems are reliable. I can expect them to be accurate most of the time.
    \item I feel safe relying on Artificial Intelligence systems for accurate information or solutions.
    \item Artificial Intelligence systems usually operate efficiently and quickly.
    \item In general, Artificial Intelligence systems can perform tasks better than novice human users.
    \item I like using Artificial Intelligence systems for decision-making.
\end{enumerate}

\noindent \textbf{Trust in People}\\

\noindent In our examination of Trust in People, we adapted the General Trust Scale from \cite{yamagishi1986provision}, originally a 6-item questionnaire designed to measure general trustworthiness and honesty among people. To align with the consistent measurement scales used throughout our study and to increase the depth of our analysis, we expanded the scale from its original 5-point Likert format to a 7-point scale ranging from 1 ("Strongly Disagree") to 7 ("Strongly Agree").

This adaptation not only facilitates consistent data collection across different constructs, but also increases the granularity with which we capture participants' levels of Trust in People. Through this adapted scale, our study aims to shed light on the comparative dynamics of Trust in People versus Trust in AI, and how these different forms of trust may differentially affect users' willingness to share data.

\begin{enumerate}
    \item Most people are basically honest.
    \item Most people are trustworthy.
    \item Most people are basically good and kind.
    \item Most people are trustful of others.
    \item I am trustful.
    \item Most people will respond in kind when they are trusted by others.
\end{enumerate}

\noindent \textbf{Privacy Concerns}\\

\noindent In assessing the construct of Privacy Concerns, our choice favored the Internet Users' Information Privacy Concerns (IUIPC) scale developed by \cite{malhotra2004internet} over the Concern for Information Privacy (CFIP) scale developed by \cite{smith1996information}. This decision was based on several key considerations. First and foremost, the IUIPC scale provides a more nuanced exploration of online Privacy Concerns that is closely aligned with the digital context of our study. Its emphasis on the perspective of Internet users as consumers directly aligns with our investigation of data sharing behaviors in online environments. While the CFIP scale provides valuable insights into individuals' concerns about organizational privacy practices and organizational responsibilities, the IUIPC scale shifts its focus to users' perceptions of fairness and justice in privacy matters, particularly when dealing with online companies \citep{gross2021validity}.

Furthermore, although integrating both the IUIPC and CFIP scales could theoretically provide a more comprehensive understanding of Privacy Concerns, such an approach was deemed impractical for our study. In particular, we were mindful of the potential for increased survey length to induce participant fatigue, which could affect response quality and engagement. Given these considerations, the decision to use only the IUIPC scale was deemed the most appropriate to capture the specific nuances of Privacy Concerns relevant to our research focus, while ensuring survey brevity and participant responsiveness.

Following \cite{malhotra2004internet} we measured each item on a 7-point Likert scale ranging from 1 ("Strongly Disagree") to 7 ("Strongly Agree"). The scores from each item are averaged to create a continuous measure of Privacy Concerns in our context, allowing for a detailed analysis of the potential effects of Privacy Concerns on data-sharing behavior.

\begin{enumerate}
    \item It usually bothers me when online companies ask me for personal information.
    \item When online companies ask me for personal information, I sometimes think twice before providing it.
    \item It bothers me to give personal information to so many online companies.
    \item I’m concerned that online companies are collecting too much personal information about me.
    \item Consumer online privacy is really a matter of consumers’ right to exercise control and autonomy over decisions about how their information is collected, used, and shared.
    \item  Consumer control of personal information lies at the heart of consumer privacy.
    \item I believe that online privacy is invaded when control is lost or unwillingly reduced as a result of a marketing transaction.
    \item Companies seeking information online should disclose the way the data are collected, processed and used.
    \item A good consumer online privacy policy should have a clear and conspicious disclosure.
    \item It is very important to me that I am aware and knowledgeable about how my personal information will be used.
\end{enumerate}

\noindent \textbf{Validation of Scales}\\

\noindent To assess the convergent and discriminant validity of our measures, we conducted a multitrait-multimethod (MTMM) analysis \citep{campbell1959convergent}. The results of the MTMM analysis showed high correlations between items within each scale (Trust in People: 0.60-0.85; Trust in AI: 0.41-0.82; Privacy Concerns: 0.57-0.78), supporting convergent validity. The correlations between items across scales were low (Trust in People and Trust in AI: 0.08-0.26; Trust in People and Privacy Concerns: -0.08-0.10; Trust in AI and Privacy Concerns: -0.12-0.09), indicating good discriminant validity.

\bibliographystyle{agsm}
\bibliography{literature}